**Preprint** Manuscript

Pattern Recognition

:
:
:
:

TEMPORAL EVOLUTION IN SYNTHETIC HANDWRITING

Cristina Carmona-Duarte , Miguel A. Ferrer , Antonio Parziale , Angelo Marcelli

# TEMPORAL EVOLUTION IN SYNTHETIC HANDWRITING

Cristina Carmona-Duarte[(1)], Miguel A. Ferrer[(1)],
Antonio Parziale[(2)], Angelo Marcelli[(2)]
[(1)] *Instituto Universitario para el Desarrollo Tecnológico y la Innovación en Universidad de Las Palmas de Gran Canaria, Spain.*
*Email: {ccarmona, mferrer}@idetic.eu*
[(2)] *Natural Computation Lab,*
*Department of Information and Electrical Engineering and Applied Mathematics,*
*University of Salerno, Italy*
*Email: {anparziale, amarcelli}@unisa.it*

## Abstract

New methods for generating synthetic handwriting images for biometric applications have recently been developed. The temporal evolution of handwriting from childhood to adulthood is usually left unexplored in these works. This paper proposes a novel methodology for including temporal evolution in a handwriting synthesizer by means of simplifying the text trajectory plan and handwriting dynamics. This is achieved through a tailored version of the kinematic theory of rapid human movements and the neuromotor inspired handwriting synthesizer. The realism of the proposed method has been evaluated by comparing the temporal evolution of real and synthetic samples both quantitatively and subjectively. The quantitative test is based on a visual perception algorithm that compares the letter variability and the number of strokes in the real and synthetic handwriting produced at different ages. In the subjective test, 30 people are asked to evaluate the perceived realism of the evolution of the synthetic handwriting.

**Highlights.**

- Method of synthesizing the temporal evolution of handwriting from childhood to adulthood.
- Synthesis of both online and offline handwriting.
- Parameters (E, $\varepsilon_D, \varepsilon_t, K_\sigma$) for dealing with synthesized handwriting evolution.
- Method for comparing temporal evolution of real and synthetic handwriting.

# 1. Introduction

Handwriting is a common tool for communication between human beings. It involves both cognitive and motor skills. Following the motor equivalence model presented in [1], the handwriting process can be divided into two stages: the effector independent stage, where the text trajectory plan is build up at cognitive level and the effector dependent stage, where the handwriting is performed by the neuromuscular system. These processes are developed during childhood by repeating patterns. Once children learn the basic patterns and are able to reproduce them, they develop their own style and evolve it up to their adulthood [2].

Aging involves some changes in handwriting characteristics. It is easy to appreciate the different writing styles between child and adult writers (see Fig. 1). In children's handwriting the pen velocity is smaller and the number of strokes greater than in the adult case [3, 4]. With aging, the handwriting tends to become slower again like that of children who are starting to write [2].

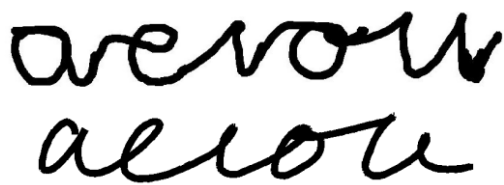

*Figure 1 Handwritten sample from a child (above) and an adult (below) writing the sequence a, e, I, o and u.*

The research on handwriting synthesis has many motivations. Among them, is to provide large handwriting corpuses to the biometric community to evaluate automatic signature verifiers or writer identifiers and to avoid legal problems on privacy [5]. It is also worth mentioning that an accurate human like synthesis mechanism could help improve the understanding of the underlying processes in human handwriting production or even answer questions related to intra and inter personal variability, as well as to help understand the variability due to different diseases, such as Parkinson's, Alzheimer's or ALS. In the future, artistic creation and CAPTCHA (Completely Automated Public Turing test to tell Computers and Humans Apart) generation may have other motivations [5,6,7,8,9,10].

There are different ways to generate synthetic handwriting. Some produce duplicates of a given handwritten sample. These duplicates can be generated by simple affine distortion or stroke wise distortion, as proposed by [11,12,13,14]. A second way of generating synthetic handwriting is the glyph-based method, which records individual letters or words from one user, applies geometric deformation to simulate a new user and joins them to create a new version of the handwriting [15, 7]. Other methods generate handwriting samples by modifying the parameters of a handwriting generation model. Handwriting models have also been developed in the frequency domain [16] or from a neuromotor perspective [17, 18]. None of the above have studied the temporal evolution of handwriting nor included handwriting evolution models in the synthesizer.

This paper is aimed at synthesizing handwriting by taking into account the *graphic maturity* of the synthetic writer for emulating its temporal evolution from childhood to adulthood. Graphic maturity is defined as the time a healthy person has been practicing his handwriting [19]. Specifically, the paper tries to answer the question: how could the writing script of writers of different graphic maturity be synthesized automatically in a common framework?

Related research has been performed on age estimation from handwriting [20] and on studying the effects of aging in signature recognition [21, 22]. It is expected that studying handwriting evolution from the synthesis point of view will deepen our understanding of the human handwriting process and its influence in designing automatic writer and signature verifiers.

As the maturity process involves both the cognitive and the motor system, the synthesizer most suitable for modelling the temporal evolution of the handwriting is the one proposed in [17], which allows actions at both cognitive and motor level. Specifically, actions at cognitive level are related to the modification of the letter engram trajectories through the spatial grid, as evident in [23]. At motor level, actions to take into account the maturity modify the Plamondon Kinematic model [24].

The model presented here is verified for three important ages: 5, 10 and adult. This is because these three ages are distinct in terms of behavioural adjustment and related to the maturation process of the neuromotor system in human beings. At the age of 5, children start to learn the motor programs required to write with pre-handwriting letter patterns. The motor programs for cursive handwriting are fully developed and integrated around age 10 but need more deliberate practice [2, 4]. By the time children reach adulthood, handwriting movements are fully mastered.

Summing up, this paper proposes a novel procedure through the use of a synthetic handwriting model to emulate the temporal evolution of real handwriting. A review of the basic handwriting synthesizer which our method relies upon is presented in Section 2, while the proposed temporal evolution model and its integration into the basic synthesizer is described in Section 3. The performance evaluation is described in Section 4. This reports the quantitative experiments based on speed profiles and stroke distributions of real and synthetic handwriting samples at different ages. It also describes surveys on subjective opinion about the temporal evolution of synthetic handwriting. Section 5 closes the paper with the conclusions.

## 2. Overview of the basic synthesizer.

The basic handwriting synthesizer is founded on the equivalence model that divides human handwriting into two steps: the working out of an action plan (effector independent) and its execution via the corresponding neuromuscular path (effector dependent). Once the action plan is learnt, most of the variability arises from the effector-dependent component [25].

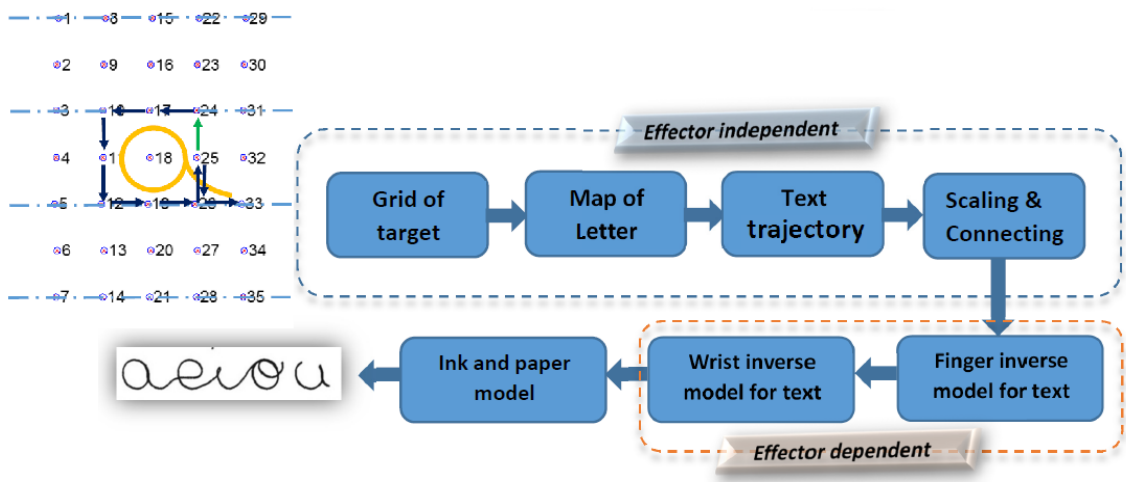

*Figure 2: Motor equivalence approach to synthetic handwriting generation and the trajectory plan for the letter 'a'.*

The synthesizer simulates the action plan through a trajectory plan which is a tessellation of a sequence of grid nodes. The neuromuscular path is calculated by the inverse model [26] as a sequence of kinematic filters that imitate the sequence of motor commands. Finally, an ink deposition model is applied. The block diagram of the synthesizer is shown in Fig. 2.

The trajectory plan is built by concatenating letter trajectory plans which describe the sequence of grid points necessary to write each letter. The letter trajectory plan defines the temporal order of the principal targets of the pen movement [27], emulating how letters are memorized [15]. An example of trajectory for letter "a" is shown in Fig. 2, where each grid point is labeled with a number. For instance, the letter trajectory plan for the letter "a" is defined as the following sequence of grid points: {25, 24, 17, 10, 11, 12, 19, 26, 25, 26 and 33}.

Once the trajectory plan is defined, an inverse model for motor control is applied to obtain a realistic human text trajectory. In short, two kinematic filters, which are heuristically related to the finger and wrist, are applied as follows:

1. the grid points of the trajectory plan are linked by straight lines and divided into strokes;
2. the finger velocity profile is estimated using the kinematic theory of rapid movements, developed in [28]. This theory shapes the velocity profile of a simple stroke with a lognormal function scaled by the variable $D$ and time-shifted by the variable $t_0$:

$$|\vec{v_j}(t; t_{0j})| = D_j \Lambda_j(t; t_{0j}, \mu_j, \sigma_j) = \frac{D_j}{\sigma_j \sqrt{2\pi}(t-t_{0j})} e^{\left(\frac{-|\ln(t-t_{0j})-\mu_j|^2}{2\sigma_j^2}\right)} \quad (1)$$

where μ and $\sigma$ are the location and scale parameters, respectively, and $j$ indicates the stroke number.

3. the finger velocity obtained with the kinematic theory is used to select the length of the inertial Kaiser filter that programs the finger control motor. The finger filter stops in each minimum of the velocity profile which could be seen as stroke limits. Conversely, the wrist moves continuously when writing and therefore the wrist inertial filter runs between penups without stopping.

The handwriting synthesizer described above is not able to simulate the learning process by which the handwriting evolves from being composed of short, imprecise, individual strokes drawn one after the other, as when a child begins to write, to the fluent movement observed in an adult, when handwriting is fully mastered. In the following section we describe the changes to the basic synthesizer to incorporate within it such temporal evolution.

## 3. Temporal evolution synthesis

Children usually start their handwriting practice using printed worksheets. These worksheets contain writing lines that guide the handwriting. At the beginning, the text trajectory plan is learned by repeating the writing on the worksheets. In a first stage, the children repeat simple traces such as small straight and then curved movement. Once children learn the basic traces, in a second stage, they combine them into complex ones (letters and numbers) by overlapping the movements. In these cases children still overwrite or copy the guide lines with short, imprecise and slow movements. Finally, once the handwriting skills are fully acquired, they are capable of selecting an ordered sequence of target points to perform fluent and personalized writing.

The synthesizer proposed in [17] is oriented towards mature and fluent handwriting because the handwriting letter shape is worked out by filtering the original trajectory plan with inertial filters that relate to adult kinematics. It does not consider a child's short and slow, unskilled movements.

Also the sigma-lognormal model used to analyze the kinematics of real handwriting movements [24] is useful in reconstructing fast and well learned movements but it is not able to fit faithfully children's dynamics and therefore obtains a poor signal-to-noise ratio in the reconstruction process [2, 4]. So a new model that enables the possibility of automatically generating dependable adult and child handwriting in terms of shape and dynamics from a trajectory plan is needed to improve the reliability and applicability of handwriting synthesizers.

### 3.1. Analysis of simple straight and curved movements.

To help model children's handwriting, two basic or simple movements are defined, as suggested in [29]: *straight movement,* which is a direct movement between two grid points namely $Q_1$ and $Q_2$, and *curved movement,* which is defined as a ballistic movement from $Q_1$ to $Q_3$ through the "via point" $Q_2$, whose curvature is modified by changing the base/height ratio of the triangle $\overrightarrow{Q_1Q_2Q_3}$. Both simple movements are illustrated in Fig. 3.

In order to model these movements, we carried out a preliminary study by asking 10 adult volunteers to use a tablet to draw straight lines and triangles, between two points, so as to obtain the curved movement or arc, as shown in Fig. 4 (left). The experiment was repeated for different scales, while maintaining the same proportions. The tablet used to collect the data was a WACOM Intuos 3 with an Intuos 3 Grip Pen, with a sampling rate of 200 Hz. The tablet has a resolution of 2540 dpi and a work surface of 304.8 mm x 228.6 mm.

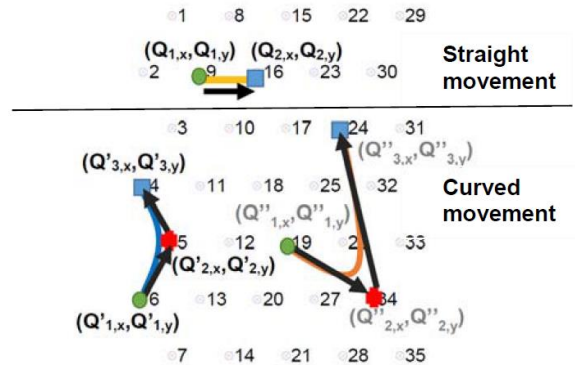

Figure 3: *Definition of the trajectory plan for a basic straight (above) and curved movements (below).*

In the case of straight lines, we observed that the speed profile starts and ends in zero velocity with a maximum between them. In the case of curved movements, we observed that the speed profile was independent of the scale but related to the angle α between the vectors $\overrightarrow{Q_1Q_2}$ and $\overrightarrow{Q_2Q_3}$. Specifically, the speed profile starts and ends with zero velocity and it displays one or two maxima depending on whether α is greater or smaller than 90º, respectively. This effect is illustrated in Fig. 4 (left), which shows how a simple movement is composed of two overlapping lognormals. The overlapping of two consecutives lognormals depends on the delay $\Delta t = t_{02} - t_{01}$ (see Equation 1). Analyzing the delay $\Delta t$, we fitted two lognormals to the velocity profile of an adult's handwriting as shown in Fig. 4 (left). As a result of the fitting, the $\mu$ and $\sigma$ values of both lognormals were set to 0 and 0.05 respectively. The result can be seen at Fig 4 (left) next to the Minimum Square Error which is obtained from superposing the adjusted lognormal variation onto the real data signal. The first lognormal starts with the curved movement. The second lognormal was delayed to fit the second peak. The conclusion drawn from this experiment is that the delay between the two lognormals is inversely proportional to the angle between $\overrightarrow{Q_1Q_2}$ and $\overrightarrow{Q_2Q_3}$. The delay (Δt) was measured for ten subjects and 17 triangles, as Fig. 4 (left) with different angles from 180 to 0 in steps of ten degrees. With these experimental values Fig. 4 (right) shows the mean normalized time increment measured for each angle and all subjects. As is also shown in Fig. 4, the experimental values can be fitted by a sigmoid curve $S(\alpha, b, -c)$, where:

$$S(\alpha, b, c) = \frac{1}{1+e^{-b(\alpha-c)}} \quad (2)$$

and where $\alpha$ is the angle in degrees, as in Fig. 4. To adjust the sigmoid, the parameters b and c were chosen to minimize the Minimum Square Error between the real averaged curve and the sigmoid curve. The experiment was carried out for values of *b* between 0 and 1 in 0.001 steps and the value of *c* between 1 and 200 in steps of 0.1. The minimum peak was found for b values between 0.06 and 0.07 and for c between 60 and 70. The values that minimize the minimum square error, with respect to the average curve of all the subjects, were for *b* = 0.06 and *c* = 65. The resulting Minimum Square Error was 0.039 between the real mean curve and the sigmoid curve.

To validate whether the two-stroke experiment generalizes to a multi-stroke letter, we asked 10 children to write the letter 'a' where there are multiple slow movements and the lognormals are sufficiently time-spaced to allow the correct measure. We decomposed into lognormals the initial part of the character (angle greater than 100º between segments) and the initial part of the terminal tail (angle near to 0º). The result can be seen in Fig. 5. The result of the initial part of the handwriting, shown at Fig. 5 (top), displays $\Delta t = 0.035$ but with smaller σ values (σ =0.01) than in the adult case because of shorter strokes. Also, in the part of the letter 'a' with smaller angles (0º) (Fig. 5, lower), the time between lognormals increases, as in adults' handwriting, increasing to $\Delta t = 0.2$. So the differences between children and adults seem not to be in the lognormal combining rules but in the action plan and in the relation of the stroke length width σ.

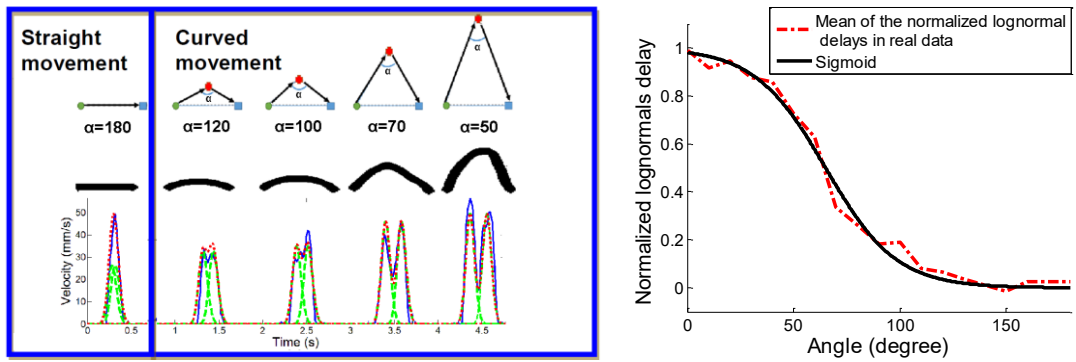

*Figure 4: Left: Real variation of the velocity profile relates to the angle (α) of the trace (triangle base of 1 cm). Lower figure: velocity profile (continuous line), individual lognormal (σ =0.05, µ =0) (dashed line) and lognormal integration (dotted line).The adult handwriting was recorded with a Wacom tablet. Right: Real measures of the time increment (dashed line) and Sigmoid approximation S(α,0.06,-65) (continuous line).*

These observations can bring about a unified framework for generating both child and adult movement in the synthesizer and therefore modelling the temporal evolution of the handwriting from childhood to adulthood, as described in the next section.

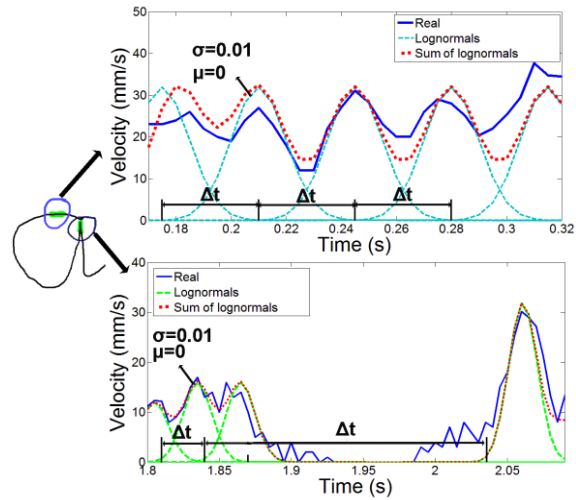

Figure 5: Corresponding child velocity profile (full line), individual lognormal (σ =0.01, µ =0) (dash line) and lognormal integration (dot line).

### 3.2. Unified framework to synthesize slow and fast movements

To allow the emulation of handwriting progress from childhood to adulthood, this section proposes a new model based on a modification to the one summarized on section 2. The new flowchart is shown in Fig. 6.

For the effector independent, the new model keeps the trajectory plan as an open polygon through a grid, but following the findings in [23], the rectangular tessellation has been updated to a hexagonal denser grid as is shown in Fig. 7.

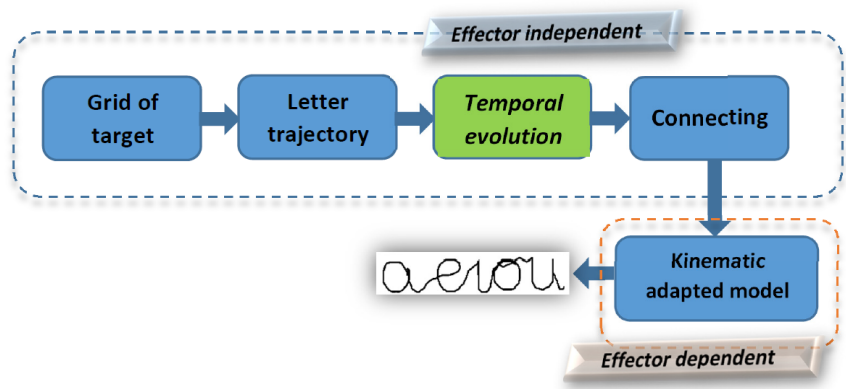

Figure 6: Scheme of the handwriting evolution model.

For the effector dependent, the new model presented in this paper considers that the speed profile is defined by one lognormal per vector $(\overrightarrow{Q_j Q_{j+1}})$ of the trajectory. The parameters $\mu_j, \sigma_j, t_{0j}$ and $D_j$ of the lognormal relative to the j-th stroke are fixed so as to work out a realistic velocity profile, that is to say, a velocity profile that could be performed by any human but not anyone in particular. These parameters are estimated as follows.

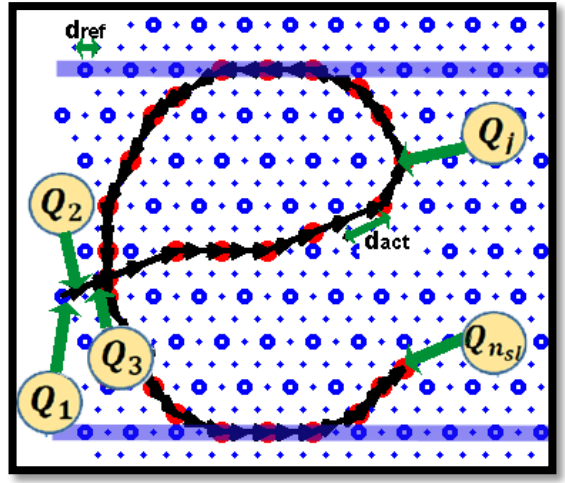

Figure 7: Grid and trajectory of letter 'e'.

From the results obtained in Section 3.1, we can assume that the scale parameter $\sigma_j$ and $\mu_j$ can be set to a constant per person as they refer to the individual's motor system and therefore his or her letter size. In this way $\sigma = \sigma_j$ and $\mu = \mu_j$. If $\sigma$ and $\mu$ are constants, the lognormals' shape is kept equal and the proportions of the letters are controlled by the amplitude of the lognormal and the time delay. Specifically, based on the experiments reported in section 3.1, $\mu$ is set to 0 and $\sigma$ is calculated as:

$$\sigma = 0.01 + K_\sigma \qquad (3)$$

where $K_\sigma$ is a writer dependent constant whose value depends on the average length of the trace. Assuming that the trace length is shorter for children than for adults, the $K_\sigma$ parameter ranges from 0 for children to 0.04 for adults, in order to reproduce the real handwriting measurements presented in section 3.1. Fig. 8 shows the effect of changing $K_\sigma$, without modifying the other parameters. It shows that as $K_\sigma$ increases so does the superposition between lognormals and the roundness of the resulting letter shape. When $K_\sigma$ reach values around 0.04, the shape is similar to the adult handwriting, but as shown by the velocity profile, the writing takes longer than in the case of $K_\sigma = 0.01$, and the lognormal width is greater. In conclusion, varying this parameter allows an evolution in the shape but not in the velocity profile.

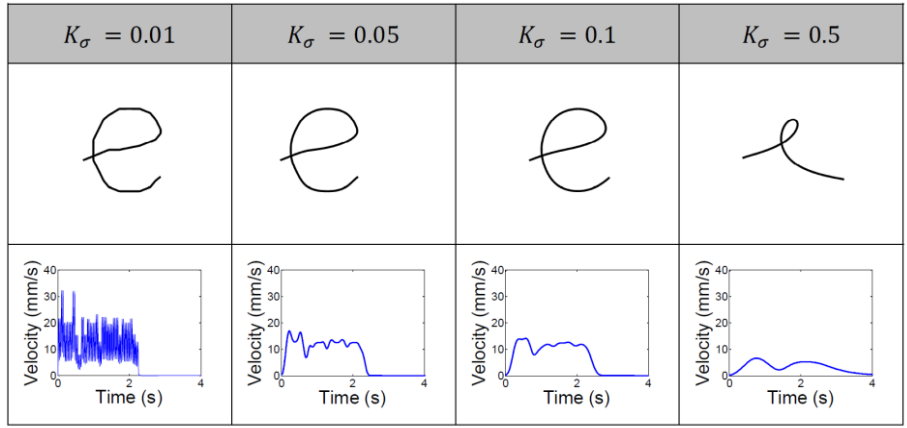

Figure 8: Result of varying $K_\sigma$, the letter shape and the velocity profile. (with $\varepsilon_t$=0, $\varepsilon_D$=0 and $K_t$=0.04)

Following the results obtained in section 3.1, the time delay between consecutive lognormals can be defined by the sum of two terms: a constant part plus an angle dependent part. Therefore the initial time $t_{0j}$ of each lognormal can expressed as:

$$\begin{cases} t_{0j} = \sum_{l=1}^{j}(K_t + \delta_l + K_\alpha S(-\alpha_j, 0.06, -65)) \\ \delta_l = N(0, \varepsilon_t), \quad l = 0,1,2, \dots, n_{sl} \end{cases} \quad (4)$$

where $\alpha_j$ is the angle between the vectors $\overrightarrow{Q_{J-1}, Q_J}$ and $\overrightarrow{Q_J, Q_{J+1}}$ defined as:

$$\alpha_j = acos\left(\frac{\langle \overrightarrow{Q_{J-1},Q_J}\rangle\langle \overrightarrow{Q_J,Q_{J+1}}\rangle}{\|\overrightarrow{Q_{J-1},Q_J}\| \|\overrightarrow{Q_J,Q_{J+1}}\|}\right), \quad j = 0,1,2, \dots, n_{sl} \quad (5)$$

where $n_{sl}$ is the number of grid points in the trajectory, $K_\alpha$ is the maximum time increment per writer (angle equal to 0°), $K_t$ is a constant which depends on the writer, letter size and average speed and $N(0, \varepsilon_t)$ is a normal random variable that emulates the time dispersion, which is inversely proportional to the skill of the writer. When $K_t$ increases, the size of the letters increases too, but if the values of $K_t$ becomes bigger than the time the lognormal is active, the superposition of the strokes is lost and the complete movement becomes a sequence of straight, independent movements (Fig. 9, $K_t$=0.06). When $K_t$ is reduced we get more superposition of the lognormals, the number of strokes decreases and the effect is that the letters are more curved and smaller (Fig. 9, $K_t$=0.005). The value of $K_t$ is set to the average time separation between lognormals in the experiment of section 3.1 and $K_\alpha$=0.2 as the maximum delay when the angle $\alpha_j$ is near to 0 as is shown in Fig. 5. In Fig. 9 we can observe how the roundness of the letter shape, the time duration of the movement and the maximum of the speed profile are inversely proportional to the $K_t$ value.

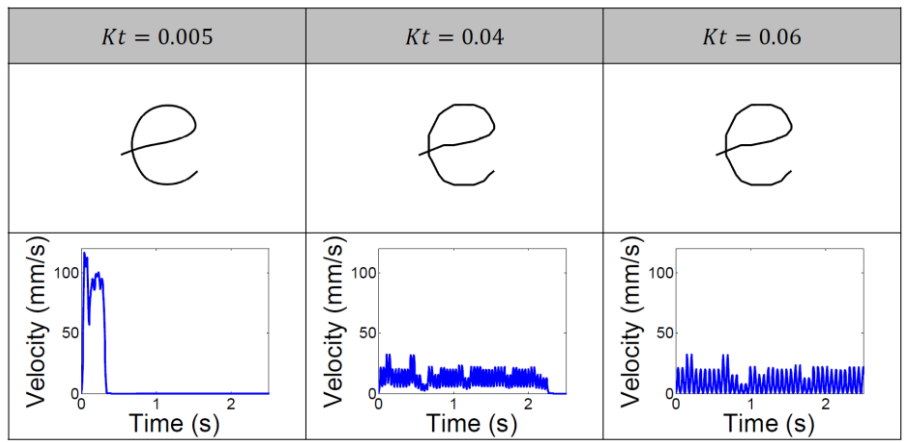

Figure 9: Result of varying $K_t$, the letter shape and the velocity profile. (with $\varepsilon_t$=0, $\varepsilon_D$=0 and $K_G$=0.01.

The parameter $\varepsilon_t$ controls the time stability. If we increase this value, the time control is less accurate resulting in deformed letters in both size and proportion. The maximum $\varepsilon_t$ is related to the variance of Δt measured from the real children's handwriting. So, in this work $\varepsilon_t = \pm 0.02$ (50 % of $K_t$) was used for children's handwriting (with deformation but keeping legibility) and $\varepsilon_t = 0$ in the case of adult handwriting synthesis. The effect of this parameter on letter shape is shown in Fig. 10: a too high value of $\varepsilon_t$ seriously affects the trajectory proportions.

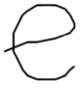

Figure 10: Result of varying $\varepsilon_t$ (with $K_t$=0.04, $\varepsilon_D$=0 and $K_\sigma$=0.01).

As the speed is proportional to the trace length, the amplitude $D_j$ of the lognormal has to be proportional to the distance between two consecutive grid points. Thus $D_j$ is given by:

$$D_j = K_D \left( \frac{d_{act,j} + N(0, \varepsilon_D)}{d_{ref}} \right) \quad j = 0,1,2, \ldots, n_{sl} \qquad (6)$$

where $d_{act,j}$ is the distance between $Q_j$ and $Q_{j+1}$, $d_{ref}$ is the distance between two grid points and $K_D$ is a constant that depends on each individual and $N(0, \varepsilon_D)$ is an aleatory value designed to give random variation in individual handwriting style. $K_D$ is fixed for a normal letter size and for a normal speed: increasing $K_D$ will increase both the letter size and the average speed. The $\varepsilon_D$ is the distance to the ideal grid point and simulates the fact that, in the learning process, the trajectory of the letter is not exactly ideal. This means that writers try to approximate the ideal letter and the result is not always satisfactory [3]. If the $\varepsilon_D$ value is too high, the letter could be deformed, illegible or not appear realistic. Based on real children's data and the observation of synthetic results, the value in this study is $\varepsilon_D = 0.3$ (30 % of error in the trajectory) and $K_D = d_{ref}$ is used in the simulation. Fig. 11 shows how the letter shape varies with this parameter and the value for which the letter becomes unreadable ($\varepsilon_D = 1$).

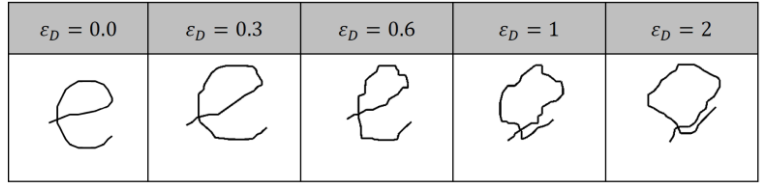

Figure 11: Results of varying $\varepsilon_t$, (with $\varepsilon_t$=0, $K_t$=0.04, and $K_\sigma$=0.01)

Once the parameters $\mu_j, \sigma_j, t_{0j}$ and $D_j$ of the lognormal of each vector $\overrightarrow{Q_j, Q_{j+1}}$ have been estimated, the speed profile $\vec{v}_j(t; t_{oj})$ is obtained by substituting their values in Equation (1). The complete synthetic speed pattern is then produced by summing up all the lognormal functions, as in [30]:

$$\overrightarrow{v_n}(t) = \sum_{j=1}^{n_{sl}} \vec{v}_j(t; t_{oj}) \tag{7}$$

Finally, to compute the spatial-temporal coordinates of the trajectory points we first compute the $x$ and $y$ components of the vector $\vec{v}_j(t)$:

$$\begin{cases} v_{x,j}(t; t_{oj}) = \text{sgn}(Q_{j+1,x} - Q_{j,x}) |\vec{v}_j(t; t_{oj})| \cos(\emptyset_j) + Q_{j,x}, j = 0,1,2, \dots, n_{sl} \\ v_x(t) = \sum_{j=1}^{n_{sl}} v_{x,j}(t; t_{oj}) \end{cases} \tag{8}$$

$$\begin{cases} v_{y,j}(t; t_{oj}) = \text{sgn}(Q_{j+1,y} - Q_{j,y}) |\vec{v}_j(t; t_{oj})| \cos(\emptyset_j) + Q_{j,y}, j = 0,1,2, \dots, n_{sl} \\ v_y(t) = \sum_{j=1}^{n_{sl}} v_{y,j}(t; t_{oj}) \end{cases} \tag{9}$$

where

$$\emptyset_j = \left| \text{atan}\left(\frac{Q_{j,y} - Q_{j+1,y}}{Q_{j,x} - Q_{j+1,x}}\right) \right| \tag{10}$$

is the angle between two successive vectors, and eventually integrate them over the movement duration:

$$x(t) = \int v_x(t) \, dt \tag{11}$$

$$y(t) = \int v_y(t) \, dt \tag{12}$$

### 3.3. Modelling temporal evolution of handwriting

As we have already noticed, children's handwriting is achieved by linking many short imprecise strokes, as the sequence of target points has not yet been learnt and therefore the letter shape is obtained by a step-by-step procedure, each step connecting two successive target points and visually locating the next one. Once the writer has learnt the letters' shape and his or her motor control has developed, the number of strokes that are required to write a text is smaller and the drawing of each stroke becomes faster [2]. This reduction in the number of strokes could be the effect of a neuromuscular velocity increment [31], which can also be interpreted as simplification of the grid, i.e. the least relevant points for letter intelligibility are suppressed [3]. As a consequence, in adult handwriting the length of the connections $\overrightarrow{Q_J, Q_{J+1}}$ increases [32].

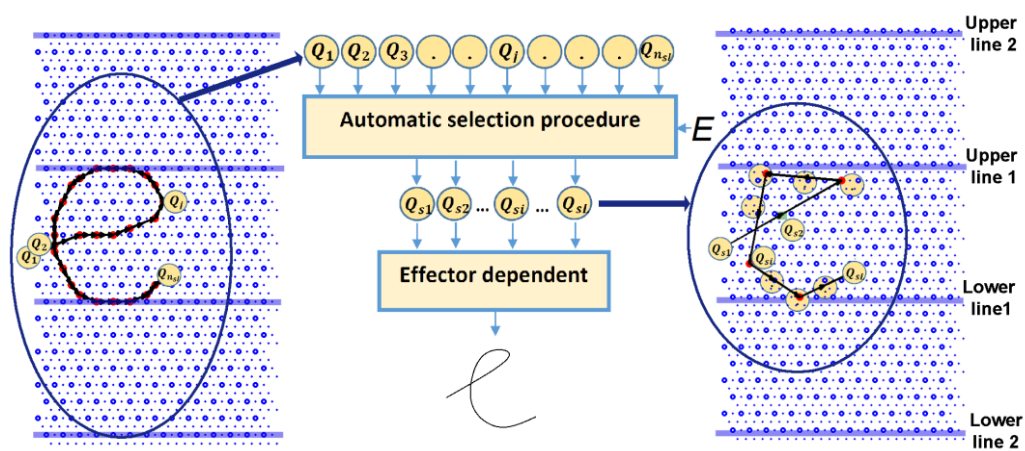

Figure 12: *Temporal model for select principal points.*

Such a behaviour is emulated in our model with a denser grid than in [17], the children's inaccuracy with the aleatory errors ($\varepsilon_D, \varepsilon_t$) and the evolution of the trajectory plan with a decreasing number of target points and strokes.

The evolution of the trajectory plan is implemented in our model by the new block entitled "*Temporal evolution*" added to the effector independent block as shown in Fig. 6: the smaller the number of grid points, the higher the level of graphic maturity of the writer. After reducing the number of grid points, the new effector dependent algorithm presented in the previous section is applied to the new trajectory vector to obtain the actual trajectory. An example of the procedure is show in Fig. 12 which shows an evolution of the letter of Fig. 7.

Accordingly, the *degree of evolution* or aging in the handwriting depends on the number of grid points selected. It can therefore be defined by the percentage *E* of points selected from the child's trajectory. Specifically, the number *L* of selected points is fixed at:

$$L = n_{sl} * \frac{E}{100} \tag{13}$$

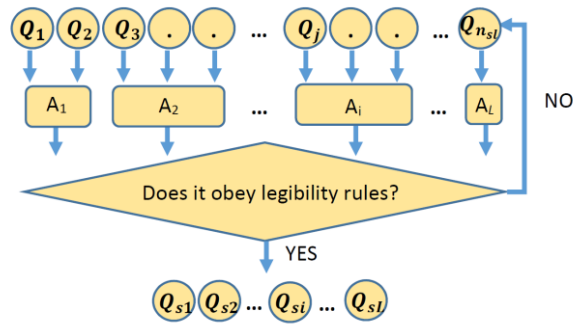

Figure 13: *Automatic selection procedure diagram.*

The value of L has to be greater than 5 to assure that both the resulting letter will be identifiable and the grid selection points will converge. So we have carefully defined the initial trajectory plan of each letter in order to secure the convergence of this algorithm, selecting the minimum *E* value as $E_{min} > 500/n_{sl}$ (0.2 in our case). The procedure for grid point selection is as follows. (see Fig.13):

1. The first two grid points are retained and the first point is randomly chosen among them.

2. The final point is also retained.

3. The remainder $n_{sl} - 3$ grid points (all of them except the two first and last one) are segmented in L clusters. Approximately $(n_{sl} - 3)/(L - 2)$ consecutive grid points belong to each cluster.

4. From each cluster we randomly select one of the grid points.

5. To guarantee legibility, we have to preserve at least one of the grid points previously located on the upper line 1 or upper line 2 or lower line 1 or lower line 2, if there are any (see Fig. 12). In case the point selection does not satisfy this constraint, we go back to step 4 and select another grid point, until the legibility rule is satisfied.

Fig. 14 shows the handwriting produced by our model by synthesizing the sequence of letters "aeiou" when the parameter E changes from 100 to 20 and the values of the model parameters $\varepsilon_D$ and $\varepsilon_t$ decrease linearly with E, from $\varepsilon_D = 0.3$ and $\varepsilon_t = 0.02$ for E=100 to $\varepsilon_D = 0$ and $\varepsilon_t = 0$ for E=20. Note that we have omitted the diacritical mark over the letter 'i'. We observe that for the value E=70 and E=80 there are no significant differences in the letter shapes because the number of points used are very similar. As the value of E decreases, the differences in the letter shapes become bigger. For a preliminary qualitative comparison, Fig. 15 shows the handwriting recorded on a tablet by children and adults. Fig. 16 also shows the comparison between the synthetic result and those of Fig. 1. In the following section the results of objective and subjective comparison are shown.

Figure 14: "aeiou" synthetic text for different values of E and decreasing $\varepsilon_D$ and $\varepsilon_t$ with E.

Figure 15: "aeiou" real text.

Figure 16: "aeiou" real and synthetic text.

## 4. Evaluation of the temporal evolution synthesizer

To evaluate the realism of the obtained results with the new handwriting synthesizer, we performed three experiments:

1. The first verifies that the evolution of the speed profile is similar in synthetic and real samples. In this way the time variation with age has been preserved.

2. The second examines whether the number of strokes and the shape variability of the synthetic handwriting are similar to real handwriting.

3. Finally, a perceptual experiment is conducted to evaluate whether a human observer can put in age order the synthetic writer's text, following the *E* aging value.

In the experiment involving human subjects, the writers' consent was given in advance. For the children, we required the consent of their parents.

### 4.1. Evaluation of the evolution with the velocity profile

For this experiment, we recruited 10 children aged 5 and 10 children aged 10 all of whom had attended school since they were two years old. We also engaged 10 educated adults. We asked each subject to write the letter 'a' on a paper grid over a WACOM Intuos 3 by an Intuos 3 Grip Pen, with a sampling rate of 200 Hz. The tablet has a resolution of 2540 dpi and a work surface of 304.8 mm x 228.6 mm. We also generated three samples of synthetic handwriting for different values of *E*: 100, 50 and 20 from a predefined prototype of each letter with a fixed number of points ($n_{sl}$) as shown Fig. 7.

Fig. 17 shows the shape and the velocity profiles of both real and synthetic handwriting. By looking at the real data, we can see that the time taken by the children to write the letter 'a' is longer than in the adult's case. Also the velocity in the children's case is almost constant with a superimposed oscillation, whereas in the adult case we can clearly distinguish three strokes or peaks.

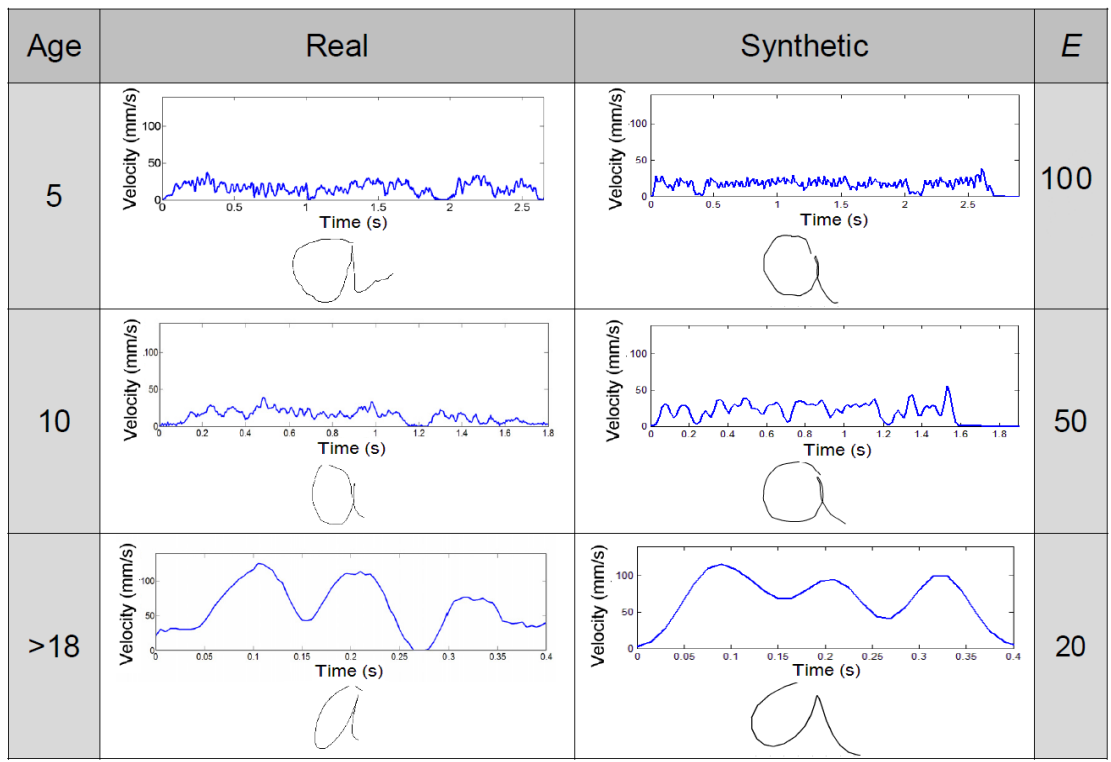

*Figure 17: Real online and offline 'a' handwriting for real (left) and synthetic (right) writers of different ages (5, 10 and greater than 18 years).*

To obtain a quantitative measure, the peaks of the velocity profile were automatically counted. In this experiment, the same subjects of the previous experiment were requested to write the word 'aeiou'. Similarly, the same text was synthetized for different values of *E*. The results are shown in Fig. 18, where we can observe how the number of peaks decrease with age along with the synthetic evolution given to the generated handwriting.

As a result, both experiments confirm that the proposed procedure of using grid selection points is able to emulate realistic progress in the evolution of handwriting from childhood to adulthood. Obviously, this curve could also be useful for comparing the differences between ages.

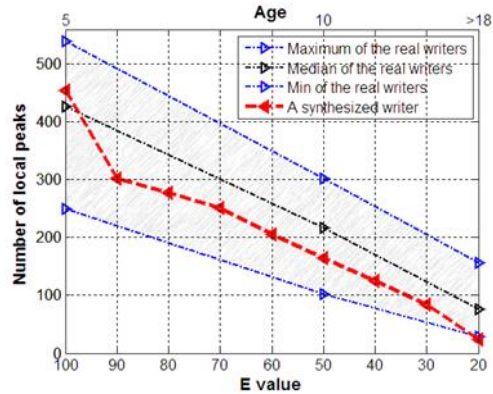

*Figure 18: (a) Number of local peaks in velocity profile of real and synthetic handwriting. Comparison of real writers with different ages (5, 10 and greater than 18 years) and a synthetic writer with different E (E from 100 to 20).*

### 4.2. Assessment of the evolution with the static images

An alternative analysis of the proposed procedure is to measure the evolution in the static images. With this aim, we conducted an analysis of the number of strokes and the handwriting similarity. For counting up the strokes in the static handwritten images we used the procedure proposed in [33]. We performed a quantitative evaluation of the similarity between two handwritten words by using the similarity measure described in [34], which is based on the Fuzzy Feature Contrast model. The method considers three sets of features to quantify the similarity between two handwritten words: 45 *zone features* relative to the position of the extracted strokes within the word itself and within the word layout; 5 *curvature features* relative to the curvature of the strokes and 4 *shape feature* relative to the word shape, as proposed in [35]. Each of these features is used as a crisp value of a linear membership function to obtain the corresponding fuzzy feature. A fuzzy feature represents the fuzziness of the presence of the feature in the handwritten word and its value is a positive real number that ranges between 0 and 1. Given a fuzzy feature vector for both the handwritten words to be compared, a score between 1 (identical letters) and 0

(completely different letters) is given for two handwriting samples by means of the relation:

$$S(a,b) = \frac{f(A \cap B)}{f(A \cap B) + \alpha f(A-B) + \beta f(B-A)} \quad (14)$$

where:

1. *a* and *b* are two images containing the two handwritten words;
2. *A* and *B* are the fuzzy feature vectors associated with *a* and *b*, respectively. The fuzzy feature vectors are computed as described above;
3. $A \cap B$ represents the intersection between the two fuzzy vectors. It is a new vector of the same size as $A$ and $B$, whose *i-th* element is equal to $min(A(i), B(i))$. It represents the extent to which a feature is present in both $a$ and $b$ and it captures the common features between $a$ and $b$;
4. $A - B$ and $B - A$ represent the complements between $A$ and $B$. They are two vectors of the same size as *A* and *B*, whose i-th elements are equal to $max(A(i) - B(i), 0)$ and $max(B(i) - A(i), 0)$, respectively. They represent the extent to which a feature is present in one of the two images but not in the other and they capture the distinctive features of '*a* in respect to *b*' and '*b* in respect to *a*';
5. $f(FV)$ is the saliency function that associates an entire feature vector $FV$ with a single number; in our implementation we choose the function $f$ as: $f(FV) = \sum_{i=1}^{54} FV_i$;
6. $\alpha$ and $\beta$ are two weights that model the imbalance of the judgment of inequality that is typically human.

For this experiment, as in the previous one, each subject was asked to write the word 'aeiou'. Also 10 images of the word 'aeiou' were synthetically generated for each of three different values of E: 100, 50 and 20.

In Fig. 19 we show the number of strokes found by this method for both real and synthetic offline images per age or E value respectively. The results are shown in the box plots in which each box represents the quartiles of each age database measure. The real data are plotted in the left (red color), and the synthetic data on the right (blue color) of each age point. If we compare this result with the previous experiment, we can observe that the number of strokes found by this procedure [33] is lower than the number of peaks in the online velocity profile. That could be possible because not all the peaks on the profile correspond to a visible variation in the handwriting image and also the error seems to be greater because of the difficulty in estimating the stroke division in the handwriting image. Anyway, as expected, the number of strokes counted in the image data decrease with increasing writer skill in both real and synthetic data. This method could also be used to compare the differences for the different ages in offline handwriting.

In order to evaluate whether a statistical difference exists between the number of strokes found in real and synthetic data, created by using the method described in [33], we performed an Anova analysis using the "*anova*" function implemented in Matlab. Two groups of strokes are considered different when the residual p-value is close to 0 and statistically similar if the p-value is greater than 0.05 [36]. The results of the Anova analysis are shown in Table 1 and Fig. 19 and they suggest that synthetic samples are second-order statistically similar to the real data for each of the ages evaluated. It seems clear that the real samples of 5 year old children, 10 year old children and adults have a similar number of strokes as those synthetic samples generated with $E$ =100, $E$ =50 and $E$ =20 respectively.

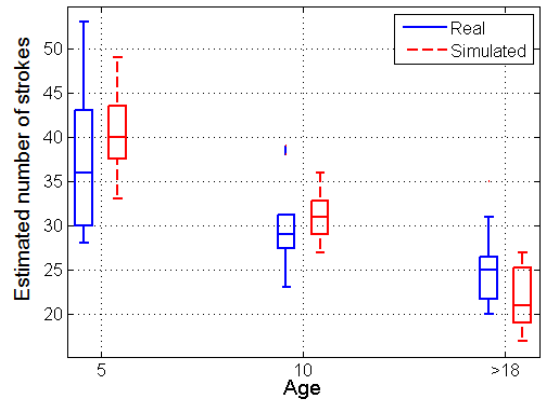

Figure 19: Offline strokes counting [33] from offline handwriting for the real handwriting for different age subjects ( 5, 10 years olds and adults) and the simulated data with different automatic selected numbers of grid points ( $E$ equal to 100, 50 and 20 ) .

Table 1: P-value results from the comparison between groups of Fig. 19 (strokes number).

|  | 5 years real | $E$ =100 synthetic | 10 years real | $E$ =50 synthetic | >18 years real | $E$ =20 synthetic |
|---|---|---|---|---|---|---|
| 5 years real | **1.0000** | 0.2796 | 0.0407 | 0.0587 | 0.0017 | 0.0001 |
| $E$ =100 synthetic | 0.2796 | **1.0000** | 0.0004 | 0.0001 | 0.0000 | 0.0000 |
| 10 years real | 0.0407 | 0.0004 | **1.0000** | 0.5134 | 0.0642 | 0.0020 |
| E =50 synthetic | 0.0587 | 0.0001 | 0.5134 | **1.0000** | 0.0054 | 0.0000 |
| >18 years real | 0.0017 | 0.0000 | 0.0642 | 0.0054 | **1.0000** | 0.1569 |
| E =20 synthetic | 0.0001 | 0.0000 | 0.0020 | 0.0000 | 0.1569 | **1.0000** |

To obtain the similarity analysis, each of the images given for a determinate age in the real case are compared with the other 6 images, thus obtaining 9 scores of similarity. The same procedure is carried out with the synthetic database. Fig. 20 shows that the distribution of similarity values of the synthetic samples for each $E$ value is similar to the distribution of the similarity values for the real samples for the different ages. Additionally, the minimum similarity score decrease when the writer age or E value increase, meaning that the shape of the letters have more variability.

As in the previous case, we have performed an Anova statistical analysis and the resulting p-values are given in Table 2. This suggests that the similarity scores in the real adults' case is similar to the similarity scores in the synthetic handwriting.

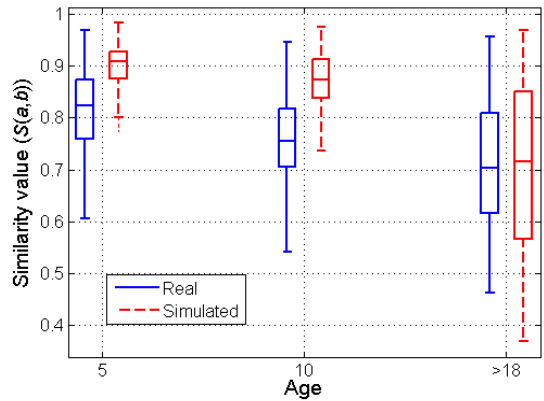

*Figure 20: Calculated similarity values (Equation 14) from offline handwriting for the real handwriting for different age subjects (5 and 10 years olds and adults) and the simulated data with different automatic selected numbers of grid points (E equal to 100, 50 and 20) .*

The difference between the minimum and the maximum of the similarity score measures the variability. With this in mind, Fig. 20 also suggests that there is less variability in the case of synthetic children's handwriting than in the real cases. In the real case, the children scores show much less variability than real adult cases. In general, there is a greater difference in variability between synthetic and real children's handwriting than between the corresponding adult cases. This is due the fact that only one case of a child's handwriting was synthesized as we explained in the previous experiment.

Table 2: p-value results from the comparison between groups of Fig. 20 (similarity value).

|  | 5 years real | E =100 synthetic | 10 years real | E =50 synthetic | >18 years real | E =20 synthetic |
|---|---|---|---|---|---|---|
| 5 years real | **1.0000** | 0.0000 | 0.0000 | 0.0000 | 0.0000 | 0.0000 |
| E =100 synthetic | 0.0000 | **1.0000** | 0.0000 | 0.0000 | 0.0000 | 0.0000 |
| 10 years real | 0.0000 | 0.0000 | **1.0000** | 0.0000 | 0.0000 | 0.0006 |
| E =50 synthetic | 0.0000 | 0.0000 | 0.0000 | **1.0000** | 0.0000 | 0.0000 |
| >18 years real | 0.0000 | 0.0000 | 0.0000 | 0.0000 | **1.0000** | 0.7385 |
| E =20 synthetic | 0.0000 | 0.0000 | 0.0006 | 0.0000 | 0.7385 | **1.0000** |

## 4.3. Perceptual evaluation

To evaluate the extent to which the modeling of temporal evolution generates samples that are perceived as produced by writers with different level of graphic maturity, we conducted a survey with 30 volunteers who were adults with a university education. The volunteers were asked to rank the handwriting from the youngest to the oldest. The handwritten samples shown to the volunteers were generated by changing only the $E$ value and were randomly ordered for each questioned volunteer. Once the words were ranked, the $E$ value of all the words with the same ranking were averaged. Fig. 21 shows the averaged result for each $E$ value. As can be seen, the respondents were able to sort out the graphic maturity with amazing reliability, except for the two first values of $E$ =80 and $E$ =70 which resulted in being indistinguishable.

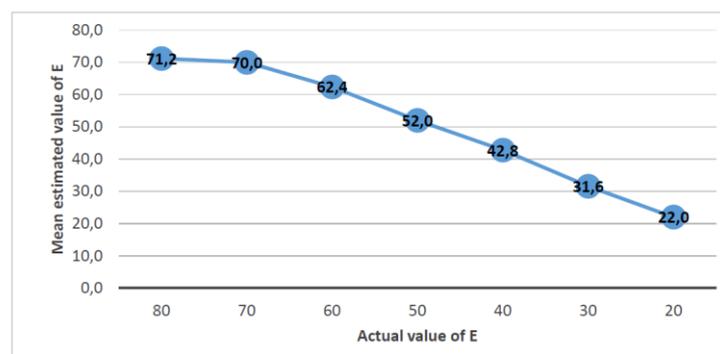

Figure 21: Mean of the E values estimated by the volunteers (Y axis) with respect of the actual E values used to generate the samples (X axis).

## 5. Conclusions and discussion

This paper proposes a novel methodology for including handwriting fluency evolution in a handwriting synthesizer. In the proposed method, changes to the graphic maturity from childhood to adulthood of synthetic handwriting are modeled with just four parameters (E, $\varepsilon_D, \varepsilon_t, K_\sigma$).

The initial handwriting synthesizer is based on the motor equivalence model which divides the human action into two steps: the effector dependent and the effector independent. The effector independent is approached by a hexagonal grid that spans the handwriting area. The action plan is defined as a sequence of grids. The effector dependent step is defined with inertial filters.

This paper redefines the effector dependent and independent step to allow the introduction of handwriting fluency progress into the synthesizer. Effectively, a lognormal shape profile is assigned to each segment of the action plan. The lognormal parameters are estimated through formulae that control the overlapping of the lognormals of the action plan, depending on the angle between consecutive segments and the writer's skill. The proposed synthesizer uses an initial dense action plan, appropriate for the early stages of handwriting development, and progressively selects a percentage $E$ of the initial set of grid points under the constraint of preserving the intelligibility of the handwriting.

Three experiments, one based on the dynamics of the synthesized handwriting, a second based on the static image of the generated trajectory and a third based on human perception of handwriting similarity, show how, by changing the percentage $E$ of selected grid points, it is certainly possible to generate synthetic handwriting which exhibits different level of graphic maturity.

This result seems to answer affirmatively the original question and confirm that it is possible to synthesize automatically handwriting of different maturity both in shape and dynamics in a common framework. The results show that when the E value decreases both the number of peaks and the intra-writer variability follow the real handwriting. The experimental results give measurable evidence of the similarity between real and synthetic handwriting in both shape and dynamics.

Comparing the results of both experiments (shape and dynamics), it should be noted that the number of strokes worked out in the dynamic domain is slightly greater than the number of strokes counted in the handwritten image. Strokes are not directly apparent in the image of a handwritten word because they are partially hidden in the trajectory as a consequence of the time superimposition process [11]. The dynamics evaluate spatio-temporal aspects of the handwritten word but the statics can evaluate only spatial aspects.

Nevertheless, in both shape and dynamics, the evolutional result of the synthetic handwriting is consistent with the well-known decreasing of the number of strokes with the practice [4]. The new method used to extract strokes from a static image [33] seems to be useful in analyzing the graphic maturity in the handwriting images.

The temporal evolution model based upon the tuning of the $\varepsilon_D$ parameter is able to produce handwritten words showing the same variation around the ideal trajectory produced by children in their early handwriting [3].

The proposed methodology could be used to generate a synthetic database for improving automatic biometric writer recognition or in CAPTCHA generation.

This study opens up the possibility of further research into handwriting evolution for elderly people. In this case, it would obviously be necessary to obtain the co-operation of a medical team who should verify the state of health of any volunteers. Moreover, similar research could be carried out to characterize degenerative diseases such as Alzheimer's, Parkinson's, essential tremor and so on.

## Acknowledgements

This study was funded by the Spanish government's MCINN TEC2016-77791-C4-1-R (AEI, FEDER) and the postdoctoral fellowship program of Universidad de Las Palmas de Gran Canaria and the Italian Department of Education, University and Research under the grant PRIN2015-HAND.

## References

[1] Lashley, K. S, The problem of serial order in behavior, In L. A. Jeffress (Ed.), Cerebral mechanisms in behavior, New York: Wiley (1951) pp. 112–131.

[2] Plamondon R., O'Reilly C., Rémi C., Duval T., The lognormal hand writer: improving, performing and declining, Front. Psychol. (2013).

[3] Grossberg, S., R.W. Paine, A neural model of corticocerebellar interactions during attentive imitation and predictive learning of sequential handwriting movements, Neural Networks 13 (2000) 999-1046.

[4] T. Duval, C. Rémi, R. Plamondon, J. Vaillant, C. O'Reilly, Combining sigma-lognormal modeling and classical features for analyzing graphomotor performances in kindergarten children, Human Movement Science, 43 (2015), pp. 183–200.

[5] M. A. Ferrer, M. Diaz-Cabrera, A. Morales, Synthetic Off-Line Signature Image Generation, 6th IAPR International Conference on Biometrics, Madrid, 2013, pp. 1 - 7.

[6] C. Ramaiah, R. Plamondonm, V. Govindaraju, A Sigma-Lognormal Model for Handwritten Text CAPTCHA Generation, in: Proceedings of the International Conference on Pattern Recognition, 2014, pp. 250-255.

[7] A.O. Thomas, A. Rusu, V. Govindaraju, Synthetic handwritten CAPTCHAs, Pattern Recognition 42 (2009) 3365–3373.

[8] Ying-Qing Xu, Heung-Yeung Shum, Jue Wang and Chenyu Wu, Learning-based system and process for synthesizing cursive handwriting, US 7227993 B2.

[9] A, Prannoy L. Roy, Method and apparatus for generating personalized handwriting, 1994, US 5327342 A.

[10] Michael R Loeb, Realistic machine-generated handwriting, 2006, WO 2006042307 A2

[11] Plamondon, R. and Guerfali, W., The generation of handwriting with delta-lognormal synergies, Biol. Cybern. 78 (1998) 119–132.

[12] Plamondon, R., et al, Recent developments in the study of rapid human movements with the kinematic theory: Applications to handwriting and signature synthesis, Pattern Recognition Letters (2012)

[13] Andreas Fischer, Réjean Plamondon, Christian O'Reilly and Yvon Savaria, Neuromuscular Representation and Synthetic Generation of Handwritten Whiteboard Notes, in: Proceedings of the International Conference on Frontiers in Handwriting Recognition, Greece, 2014, pp. 222-227.

[14] M. Diaz-Cabrera, A. Fischer, R. Plamondon, M. A. Ferrer, Towards an Automatic On-Line Signature Verifier Using Only One Reference Per Signer, 13th International Conference on Document Analysis and Recognition (ICDAR 2015), (2015), pp. 631-635.

[15] A.Lin, L.Wang, Style-preserving English handwriting synthesis, Pattern Recognition 40 (2007) 2097–2109.

[16] J. Galbally, R. Plamondon, J. Fierrez, J. Ortega-Garcia, Synthetic on-line signature generation. Part I: Methodology and algorithms, Pattern Recognition 45 (2012) 2610-2621.

[17] Ferrer, M.A., Diaz-Cabrera, M., Morales, A., Static Signature Synthesis: A Neuromotor Inspired Approach for Biometrics, IEEE Transactions on Pattern Analysis and Machine Intelligence 37 (2015) 667-680.

[18] Ferrer M. A., Diaz M., Carmona C., and Morales A.: A behavioral handwriting model for static and dynamic signature synthesis, IEEE Trans. Pattern Anal. Mach. Intell., 2016, In press.

[19] Ron Morris, Ron N. Morris, Forensic Handwriting Identification: Fundamental Concepts and Principles, Academic Press, 2000.

[20] Somaya Al-Máadeed , Abdelaali Hassaïne, Automatic prediction of age, gender, and nationality in offline handwriting, in: Proceedings of the EURASIP J. Image and Video Processing, 2014.

[21] Lanitis A, A survey of the effects of aging on biometric identity verification, International Journal of Biometrics 2, (2010) 34–52.

[22] J. Galbally, M. Martinez-Diaz and J. Fierrez, Aging in Biometrics: An Experimental Analysis on On-Line Signature, PLOS ONE 8 (2013) , e69897.

[23] Hafting, T., Fyhn, M., Molden, S., Moser, M. B. and Moser, E. I. Microstructure of a spatial map in the entorhinal cortex, Nature 436 (2005) 801-806.

[24] Plamondon, R.. A kinematic theory of rapid human movements. Part I: Movement representation and generation. Biological Cybernetics 72 (1995) 295–307.

[25] A. Marcelli, A. Parziale and R. Senatore, Some Observations on Handwriting from a Motor Learning Perspective, in: Proceedings of the Workshop on Automated Forensic Handwriting Analysis, Washington DC, 2013, pp. 6-10.

[26] M. Kawato, Internal models for motor control and trajectory planning, Current Opinion in Neurobiology 9 (1999) 718–727.

[27] Bernstein N, The coordination and regulation of movements. Pergamon, Oxford, 1967.

[28] O'Reilly C., Plamondon R., Development of a Sigma-Lognormal representation for on-line signatures. Pattern Recognition 42 (2009) 3324–3327.

[29] T. Flash and N. Hogan, The Coordination of Arm Movements: An Experimentally Confirmed Mathematical Model, The Journal of Neuroscience 5 (1985) 1688-1703.

[30] M. Djioua, R. Plamondon, A new algorithm and system for the characterization of handwriting strokes with delta-lognormal parameters, IEEE Transactions on Pattern Analysis and Machine Intelligence, 31 (2009) 2060–2072.

[31] Plamondon R and Privitera C, A neural model for generating and learning a rapid movement sequence, Biol Cybern 74 (1995) 117–130.

[32] McNaughton, B. L., Battaglia, F. P., Jensen, O., Moser, E. I., Moser, M. B., Path integration and the neural basis of the "cognitive map." , Nat. Rev. Neurosci. 7 (2006) 663–678.

[33] Cordella, L. P.; De Stefano, C.; Marcelli, A.; Santoro, A., A New Graph Search Algorithm for Writing Order Recovery, In: Proceedings IEEE Computer Society ICPR'10 , 2010, pp. 1899.

[34] A. Parziale, S. Davino, A. Marcelli. An algorithm based on visual perception for handwriting comparison. Céline Rémi; Lionel Prévost; Eric Anquetil. 17th Biennial Conference of the International Graphonomics Society, 2015.

[35] Powalka, R.K, & al, Word Shape Analysis for a Hybrid Recognition System, Pattern Recognition 30 (1997), pp. 421.

[36] Hogg, R. V., J. Ledolter, Engineering Statistics. New York: MacMillan, 1987.